\documentclass[10pt,journal,compsoc]{IEEEtran}

\usepackage{algorithm}
\usepackage{algpseudocode}
\usepackage{cite}
\usepackage{amsmath,amssymb,amsfonts}
\usepackage{textcomp}
\usepackage{booktabs}
\usepackage{multirow}
\usepackage{graphicx}
\usepackage{diagbox}
\usepackage{subfigure}
\usepackage{color}

\usepackage[pagebackref,breaklinks,colorlinks]{hyperref}

\ifCLASSINFOpdf
\else
\fi
\begin{document}
%
\title{Real-time Online Multi-Object \\ Tracking in Compressed Domain}
%
%
%
%

\author{Qiankun Liu, Bin Liu, Yue Wu, Weihai Li, Nenghai Yu\\
    \IEEEcompsocitemizethanks{
    	\IEEEcompsocthanksitem Qiankun Liu, Bin Liu, Weihai Li and Nenghai Yu are with School of Information Science and Technology, University of Science and Technology of China (Email: liuqk3@mail.ustc.edu.cn; \{flowice, whli, ynh\}@ustc.edu.cn).
        \IEEEcompsocthanksitem Yue Wu is with Alibaba Group (Email: matthew.wy@alibaba-inc.com).
        \IEEEcompsocthanksitem This work was done in February, 2019.
    }
}

\markboth{\space}%
{Liu \MakeLowercase{\textit{et al.}}: Real-time Online Multi-Object Tracking in Compressed Domain}
%



\IEEEtitleabstractindextext{%
\begin{abstract}
Recent online Multi-Object Tracking (MOT) methods have achieved desirable tracking performance.
However, the tracking speed of most existing methods is rather slow.
Inspired from the fact that the adjacent frames are highly relevant and redundant,
we divide the frames into key and non-key frames respectively and track objects in the compressed domain.
For the key frames, the RGB images are restored for detection and data association.
To make data association more reliable, an appearance Convolutional Neural Network (CNN) which can be jointly trained with the detector is proposed. 
For the non-key frames, the objects are directly propagated by a tracking CNN based on the motion information provided in the compressed domain.
Compared with the state-of-the-art online MOT methods,
our tracker is about $6\times$ faster while maintaining a comparable tracking performance.
\end{abstract}

}

\maketitle

\IEEEdisplaynontitleabstractindextext

%
\IEEEpeerreviewmaketitle

\IEEEraisesectionheading{\section{Introduction}\label{sec:introduction}}
\IEEEPARstart{M}{ulti}-Object Tracking (MOT) is an important computer vision task which aims to estimate the trajectories of interested objects and maintain their identities across frames.
It has various applications that with real-time and online requirements, such as autonomous driving and robot navigation.
However, real-time online MOT still remains a challenging task. 

\begin{figure}[tp]
    \centering
    \includegraphics[width=0.99\linewidth]{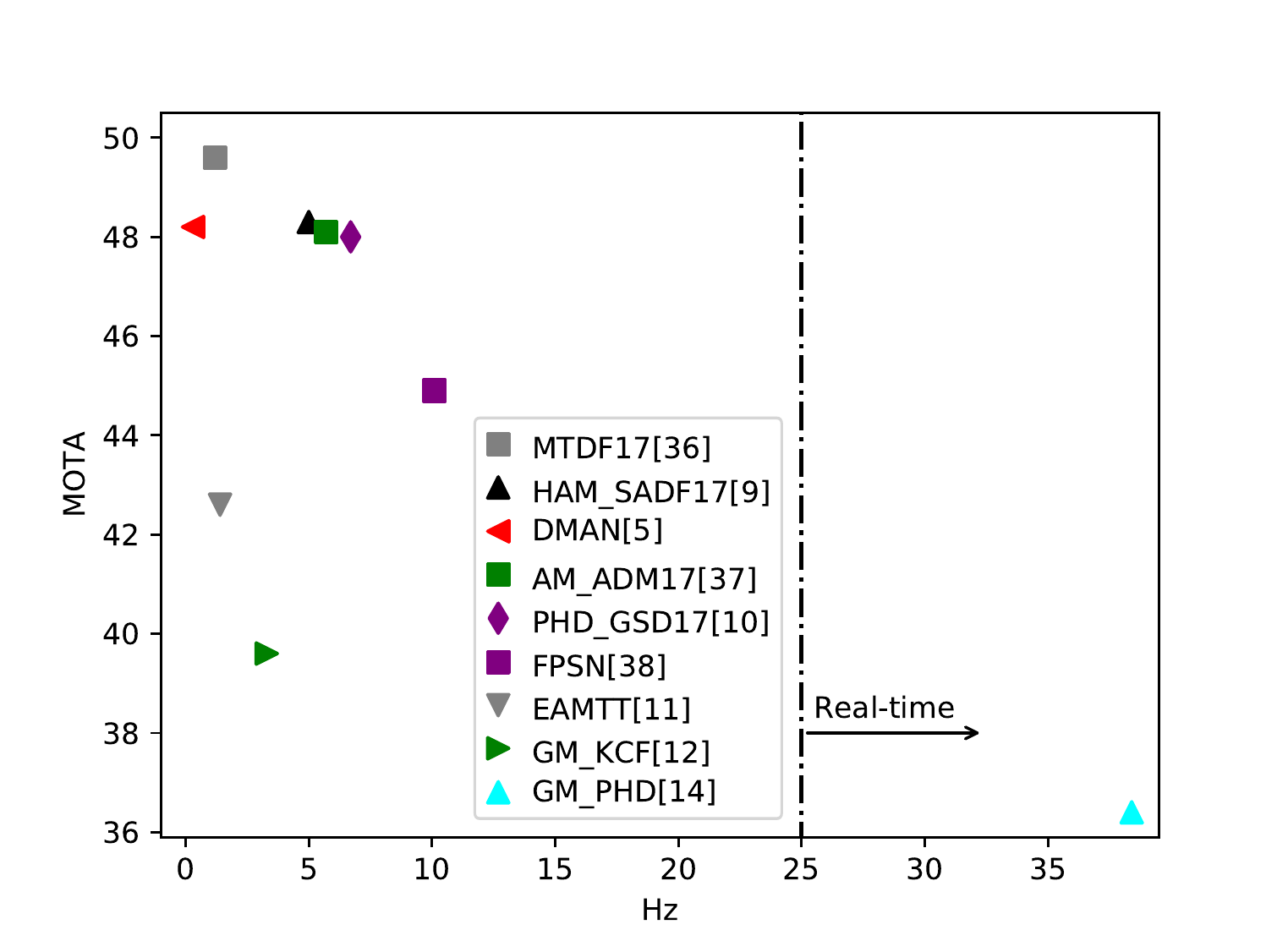}
    \caption{Performance-speed of some online trackers on MOT17 test split.
    Vertical axis: Multi-Object Tracking Accuracy (MOTA) \cite{bernardin2008evaluating}.
    Horizontal axis: the number of frames that the tacker can process in one second, i.e., the frequency.
    Better trackers lie in top-right of the figure.}
    \label{fig:motivation_mota_hz}
\end{figure}

Driven by advances in object detection, tracking-by-detection has become a popular strategy for MOT.
Most existing methods focus on designing a complicate approach to tackle MOT in a data association manner. 
These methods can be divided into two categories: offline and online methods.
The offline methods \cite{zhang2008global, tang2017multiple, yang2012online} usually use future frames to track objects, which makes them impractical for casual applications.
On the contrary, the online methods
~\cite{bae2018confidence, zhu2018online, wojke2017simple, chu2017online, ujiie2018interpolation, yoon2018online, fu2018particle, sanchez2016online, kutschbach2017sequential, yu2016poi, eiselein2012real}
track objects based on the past and current frames
and have achieved desirable performance, as shown in Figure~\ref{fig:motivation_mota_hz}.
However, 
detection and data association are performed frame-by-frame in these online methods to ensure a good tracking performance,
which is time-consuming and makes them unable to be applied in real-time applications. 
In order to re-recognize objects when occlusion happens, appearance affinity is commonly used in data association process
\cite{bae2018confidence, kieritz2016online, zhu2018online, wojke2017simple, chu2017online, yu2016poi}.
However, the appearance models utilized in these methods are independent from the detector and are therefore potentially sub-optimal.

In practical applications, videos are usually captured, stored and transmitted in the compressed domain.
Implementations for some video tasks in the compressed domain are necessary and more suitable.
Firstly, implementations in the compressed domain can take lower computational cost since not all frames need to be restored into RGB images
(the RGB images in this paper denotes the regular colorful or gray images, which is used to distinguish them from the frames in the compressed domain).
Secondly, the motion information is readily provided in the compressed domain which is helpful for video tasks.
A few works, such as tracking~\cite{ujiie2018interpolation, alvar2018mv},
video object detection \cite{wang2019detection} and action recognition \cite{wu2018compressed}, have been done in the compressed domain.
Generally, there are two goals of the implementations in the compressed domain:
(1) Feature propagation \cite{wang2019detection}.
Features are only extracted from the key frames which are restored into RGB images, and then propagated to the non-key frames.
Computational cost can be saved since the frequency of feature extraction is reduced and the feature propagation is much more efficient than feature extraction.
(2) Motion cues extraction~\cite{ujiie2018interpolation, alvar2018mv, wu2018compressed}.
The motion cues of objects are directly extracted from the motion information (motion vectors and residuals, more specifically) without access to the RGB images, which are further used to handle the task.

Existing compressed domain based tracking algorithms focus on the motion cues extraction from the pixel level \cite{alvar2018mv} or the bounding-box level \cite{ujiie2018interpolation}.
For the pixel level, each pixel that locates in the bounding-box is shifted separately based on the motion vectors (MVs),
then the smallest axis-aligned rectangle that includes all the shifted pixels is selected as the new bounding-box.
For the bounding-box level, MVs that locate in the bounding-box are averaged to get the displacement, then the bounding-box is shifted.
However, the scale variation can not be handled.

In this paper, we focus on real-time online multi-object tracking.
To this end, we propose an Online MOT Tracker in \emph{Compressed Domain} (OTCD).
Since the adjacent frames are highly relevant, it is redundant to perform detection and data association in all frames.
Motivated by this, we divide the frames into key and non-key frames respectively.
For the key frames, detection and data association are performed.
An \emph{Appearance CNN} (A-CNN) which shares features with the detector is designed to assist data association,
and it can be jointly trained with the detector.
Note that RGB images are restored for the key frames since both detection and appearance feature extraction need to be performed on the RGB images.
For the non-key frames, objects are directly propagated based on the motion cues which are extracted from the MVs and residuals by a \emph{Tracking CNN} (T-CNN).
Owning to the sparsity of key frames and the share of features between A-CNN and detector,
our tracker achieves a great boost in tracking speed with little performance degradation.

To sum up, our contributions are as follows:
\begin{itemize}
\item We develop an online unified MOT tracker to track objects in compressed domain for real-time applications. 
\item We propose an appearance CNN to assist data association. The joint training of appearance CNN and detector
  helps to further promote the performance of our method.
\item We propose a tracking CNN to propagate objects through non-key frames while maintaining their identities without detection and data association,
  which accelerates our tracker greatly with little performance degradation.
\end{itemize}
 The rest of this paper is organized as follows. Section~\ref{section_related_work} reviews the related work. Section~\ref{section_method} introduces the proposed tracker in detail and section~\ref{section_experiments} represents experimental results. Finally, section~\ref{section_conclusion} makes a conclusion on our work in this paper.

\section{Related work}
\label{section_related_work}
In this section, we provide a brief overview about the usage of appearance features in MOT and the works implemented in the compressed domain.

\noindent\textbf{Appearance Features in MOT.}
Appearance features can be used to improve tracking performance in crowded scenario where occlusion often happens.
And various appearance features have been used in MOT,
such as histogram of gradients \cite{kuo2010multi, wang2014tracklet},
color histogram \cite{kuo2010multi} and integral channel features \cite{kieritz2016online}.
Recently, the powerful deep features extracted by CNN have been introduced to MOT \cite{bae2018confidence, kieritz2016online, zhu2018online, wojke2017simple, chu2017online, yu2016poi}.
Bae \emph{et al.} \cite{bae2018confidence} proposed a deep appearance learning method to learn a discriminative appearance model in an online manner.
Kieritz \emph{et al.}~\cite{kieritz2016online} designed an appearance model which was incrementally trained online for each object.
Both~\cite{bae2018confidence} and~\cite{ kieritz2016online} need to collect the training samples online, which is time-consuming.
Chu \emph{et al.} \cite{chu2017online} utilized single object tracker for MOT based on appearance features, but the online learned target-specific CNN layers need to be preserved for each target.
Instead of learning the appearance model online, some researchers trained an appearance model offline \cite{zhu2018online, wojke2017simple, yoon2018online, yu2016poi},
which can be used as a function to measure the affinity between different features while tracking online.
Zhu \emph{et al.} \cite{zhu2018online} trained a spatial attention CNN which can focus on matching patterns of input image patches.
The works in \cite{wojke2017simple, yoon2018online, yu2016poi} trained appearance models using person re-identification dataset and achieved great improvements.

The aforementioned methods separated data association from detection,
and the utilized appearance models were isolated from the detector.
Our work focuses on designing a compact appearance model, which shares features 
and can be jointly trained with the  detector.

\noindent\textbf{Works in the Compressed Domain.}
In order to use the motion information provided in the compressed domain freely,
a few works \cite{ujiie2018interpolation, alvar2018mv, wang2019detection, wu2018compressed} have been done in the compressed domain.
Ujiie \emph{et al.} \cite{ujiie2018interpolation} interpolated the bounding-boxes of objects in the bounding-box level
for some frames to avoid detection and data association.
Alvar \emph{et al.} \cite{alvar2018mv} constructed approximate bounding-box of the target object in the pixel level
based on the bounding-box in previous frame for single object tracking, but each frame needs to be restored into RGB image for detection.
The work in \cite{wang2019detection} fed MVs and residuals to a network to propagate the features across frames for object detection.
However, the frames are processed in a batch manner which is inapplicable for online tasks.
Wu \emph{et al.} \cite{wu2018compressed} recognized different actions in the compressed domain and achieved great performance.
Nevertheless, the MVs and residuals are traced back to the reference frame
and accumulated on the way, which augments the computational cost.

Among these methods, the work in \cite{ujiie2018interpolation} is the most related to our work.
However, the work in \cite{ujiie2018interpolation} cannot handle the scale variations of bounding-boxes
since MVs are simply used (by averaging MVs that locates in the bounding-boxes) to predict the displacements of objects. 
While our method utilizes a tracking CNN to predict the velocities of objects based on MVs and residuals,
in which the scale variation of bounding-boxes are considered.

\section{Method}
\label{section_method}
Frames in a compressed video are divided into Group of Pictures (GOP),
and there are three types of frame generally: I-frame (intra-coded frame), P-frame (predictive frame) and B-frame (bi-directional frame).
Among these three types of frame, I-frame can be treated as a regular RGB image, while P-frame and B-frame are encoded with MVs and residuals.
The main difference between P-frame and B-frame is that P-frame is encoded in a predictive manner, while B-frame is encoded in a bi-directional manner.
We share the same assumption with \cite{ujiie2018interpolation, alvar2018mv, wang2019detection, wu2018compressed}
that the compressed video only contains I-frame and P-frame for simplicity.
The reason is that the B-frames, which are encoded in a bi-directional manner, require special handling since we focus on online tracking,
and we leave this for the future work.
In this paper, we track objects in raw MPEG-4 videos which have an I-frame before every 11 P-frames.
However, the proposed method is applicable for different compression techniques,
such as MPEG-2 \cite{union1994generic} and H.264 \cite{wiegand2003draft}. The reason is that different compression techniques
usually use motion vectors and residuals for frame compression,
thus the frames in the videos can be easily divided into key and non-key frames.

\begin{figure}[tp]
    \centering
    \includegraphics[width=1.0\linewidth]{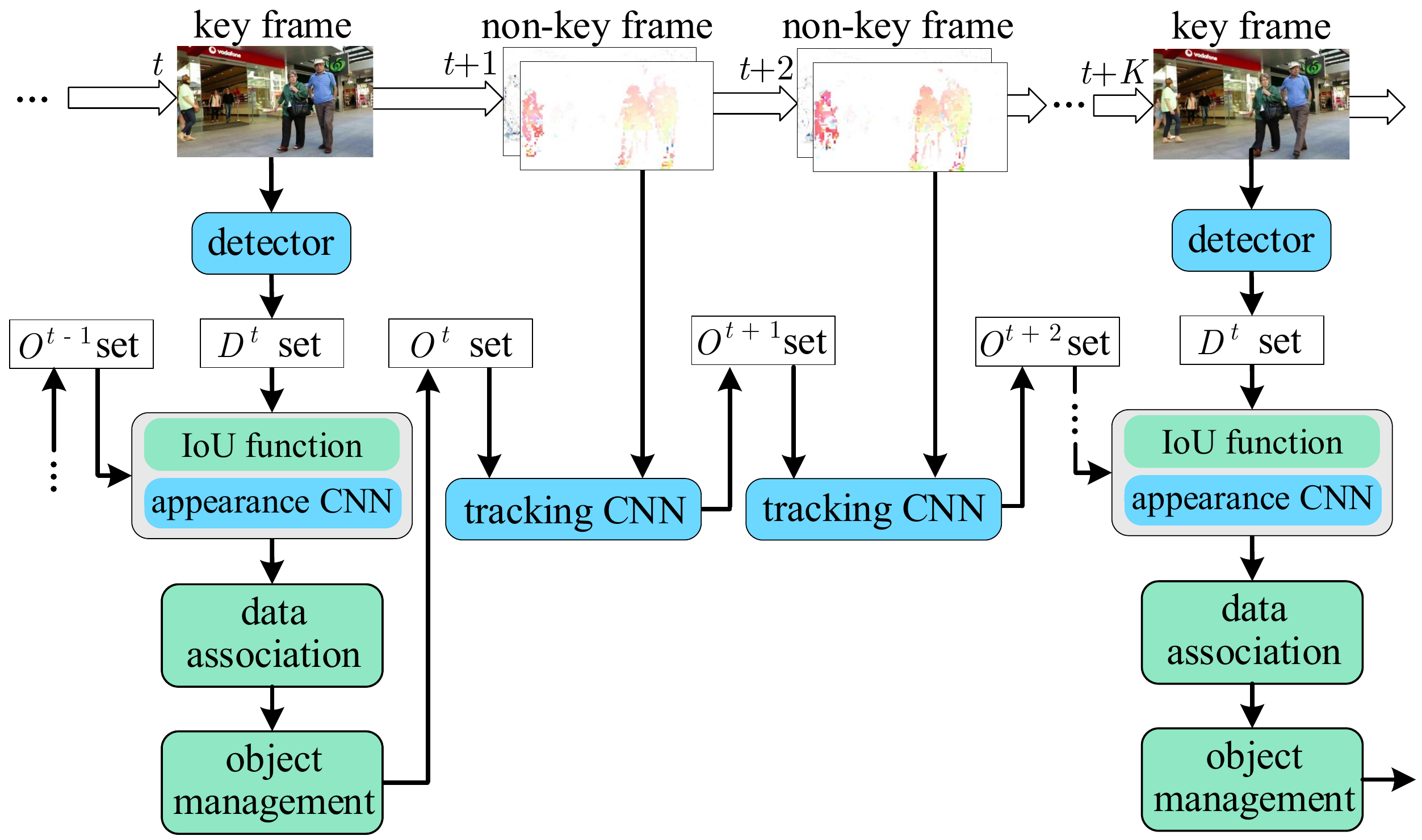}
    \caption{Overview of the proposed tracker. MVs are plotted in HSV color space.
    $O^t$ is the set of objects in frame $t$, $D^t$ is the set of detections in key frame $t$.
    Detection, data association and object management are only performed in key frames.
    Objects are propagated through non-key frames by tracking CNN while maintaining their identities.}
    \label{fig:tracker_overview}
\end{figure}

\subsection{Overview}
\label{subsection_overview}
The overview of the proposed tracker OTCD is shown in Figure \ref{fig:tracker_overview}.
The frames are divided into key and non-key frames in OTCD.
Suppose there is a key frame every $K$ consecutive frames.
Since MVs and residuals are required for the non-key frames
and there are no MVs or residuals for I-frames in the compressed domain,
I-frames are always regarded as the key-frames,  which means $K$ should be a factor of GOP size.

Let $O^{t} = \{o_i^t\}_{i=1}^{I_t}$ and $D^t = \{d_j^t\}_{j=1}^{J_t}$ denote the sets of objects and detections in frame $t$ respectively.
Note that $D^t$ is defined for the key frames only.
For a key frame at time $t$, the RGB image is restored and fed into a detector, which produces a set of detections $D^t$.
The objects in $O^{t-1}$ from last frame are associated with the detections in $D^t$.
The data association is solved by Hungarian algorithm based on the Intersection-over-Union (IoU) between bounding-boxes and the appearance affinity obtained by A-CNN.
After then the birth and death of objects are managed.
For each non-key frame, the corresponding MVs and residuals are fed into T-CNN to propagate objects from the previous frame to current frame.

Before introducing the method in detail, we first introduce the representations of objects and detections.
The $j$-th detection in $D^t$ is denoted by a tuple $d_j^t = (b_j^t, f_j^t)$, 
where $b_j^t = (x_j^t, y_j^t, w_j^t, h_j^t)$ is the bounding-box represented by the center coordinate, width and height.
$f_j^t \in \mathbb{R}^{m \times m \times c}$ is the appearance feature cropped by RoIAlign \cite{he2017mask} from the feature map provided by detector,
where $m$ and $c$ are the spatial size and the number of channels respectively.
As for objects, three states $\{Tentative, Confirmed, Deleted\}$ are defined to handle the birth and death of objects, which are denoted by $\{s_T, s_C, s_D\}$ for simplicity.
The $i$-th object in $O^t$ is denoted as $o_i^t = (b_i^t, s_i^t, F_i)$,
where $b_i^t = (x_i^t, y_i^t, w_i^t, h_i^t)$ is the bounding-box, $s_i^t \in \{s_T, s_C, s_D\}$ is the state.
And $F_i = \{f_i^{t-\tau}\}_{\tau=0}^{l_{f} - 1}$ is the set of appearance features collected in the history,
where $l_{f}$ is the maximum number of appearance features.

\subsection{Tracking in Key Frames}
\label{subsection_tracking_in_key_frames}
The tracking in key frames follows the footprint of per-frame approaches,
including detection, data association and object management.

\subsubsection{Detector}
The detector is responsible for the detection of interested objects (pedestrian, particularly) and feature extraction.
The R-FCN \cite{dai2016r} with ResNet-101 \cite{he2016deep} is used in our work directly since detection is beyond the scope of this paper.
We take pedestrian as the foreground and others as the background for detection.
The appearance features for detections are cropped from the feature map provided by the last convolutional layer on the conv4 stage of
the backbone of detector.

\begin{figure}[tp]
\centering
    \includegraphics[width=1.0\linewidth]{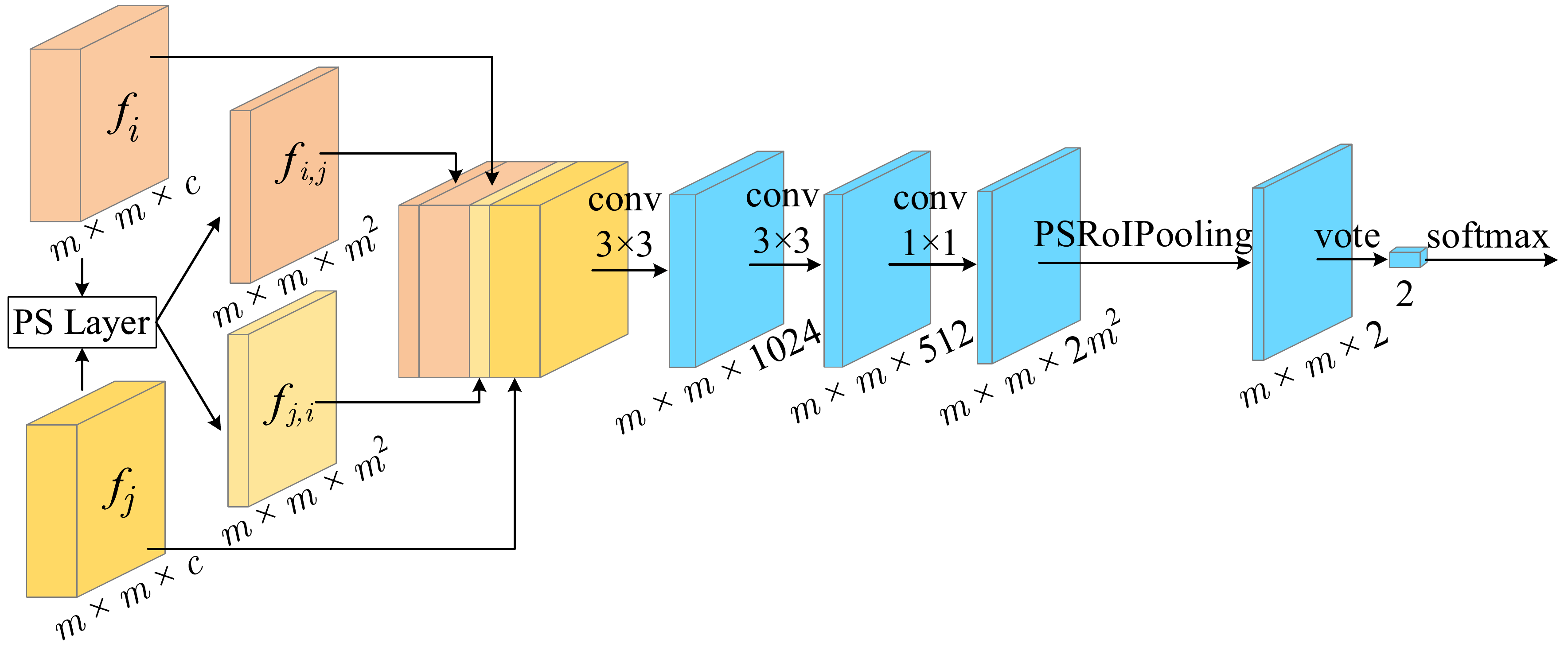}
    \caption{Architecture of A-CNN. Each convolution layer is followed by a ReLU function.
    A position sensitive (PS) layer is introduced to exploit the position information between input features.}

\label{fig:appearance_cnn}
\end{figure}

\subsubsection{Appearance CNN}
Given the appearance feature $f_j^t$ of the detection $d_j^t$ and one appearance feature $f_i^{t-\tau} \in F_i$ of the object $o_i^{t-1}$,
the probability $p_{i,j}^\tau$ of these two features belonging to the same object is used as the affinity between these two features:
\begin{equation}
p_{i,j}^{\tau} = \mathcal{N}_A(f_i^{t-\tau}, f_j^t),
\label{equation_a_cnn}
\end{equation}
where $\mathcal{N}_A(\cdot, \cdot)$ denotes A-CNN.
As shown in Figure \ref{fig:appearance_cnn}, A-CNN is a binary classifier in fact.
Note that the appearance features $f_j^t$ and $f_i^{t-\tau}$ are cropped from the feature map provided by the conv4 stage of the backbone of detector,
and no feature extraction (except the three convolutional layers) is performed within A-CNN,
which means A-CNN shares features with the detector.
The superscripts are omitted for simplicity in the following.

\noindent\textbf{Position-Sensitive Layer.}
Given two features $f_i, f_j \in \mathbb{R}^{m \times m \times c}$, 
we try to obtain the affinity between these two features.
The $L^2$-normalization is first applied to $f_i$ and $f_j$ along the channel dimension,
which produces the corresponding normalized features $f'_{i}$ and $f'_{j}$.
Then two position-sensitive feature maps $f_{i, j}, f_{j, i} \in \mathbb{R}^{m \times m \times m^2}$ are produced respectively:
\begin{equation}
    f_{i,j} =  f'_i \otimes \mathcal{T}(f'_j), \quad
    f_{j,i} =  f'_j \otimes \mathcal{T}(f'_i),
\label{equation_position_matrix_multuplication}
\end{equation}
where $\otimes$ denotes the matrix multiplication,
and $\mathcal{T}(\cdot)$ is the function that
reshapes and transposes a $3$-D feature map in $\mathbb{R}^{m \times m \times c}$ to a $2$-D feature map in $\mathbb{R}^{c \times m^2}$.
Each column in $\mathcal{T}(f'_i)$ corresponds to a feature vector that locates in one spatial position in $f'_i$.
Finally, the four feature maps $f_{i, j}$, $f_{i}$, $f_{j, i}$ and $f_{j}$ are concatenated together.
Despite the values in $f_{i,j}$ and $f_{j,i}$ are the same, the distributions of them are different.
We keep both of them to preserve more position information.

The intuition of the Position-Sensitive (PS) layer is that we assume the features extracted from the same patches in different RGB images should be the same,
but the patches may not well aligned due to the inaccurate detection, occlusion and pose change.
The corresponding features in $f'_{i}$ and $f'_{j}$ may locate in different spatial positions.
Hence, it is necessary to compare the feature vector from one spatial position in $f'_{i}$ with the feature vectors from all spatial positions in $f'_{j}$,
which produces a single channel feature map in $f_{j,i}$.

\noindent\textbf{Training of A-CNN.}
During the training process, each training sample contains two appearance features cropped from
the feature map provided by the detector.
The corresponding label is set to 0 (these two appearance features belong to different objects) or 1 (these two appearance features belong to the same object).

A-CNN is trained by the cross-entropy loss. Let $L_A$ and $L_D$ be the loss of A-CNN
and detector respectively.
Then A-CNN and the detector can be jointly trained via a multi-task loss $L = L_D + \lambda L_A$,
where $\lambda$ is the weight to balance the loss.

\begin{figure}[tp]
\centering
    \includegraphics[width=1.0\linewidth]{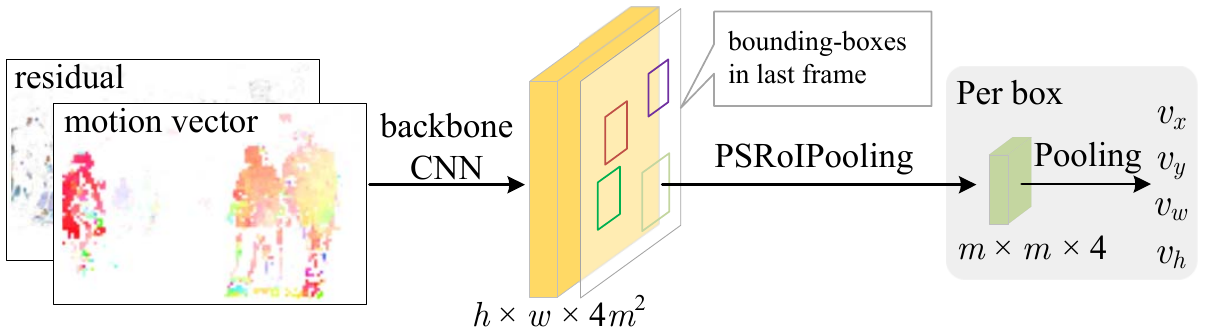}
    \caption{The Velocity prediction.
    The velocity of each un-deleted object is estimated by PSRoIPooling based on the bounding-box in last frame.
    The number of bounding-boxes in current frame is the same with that in last frame.}
\label{fig:tracking_cnn_overview}
\end{figure}

\subsubsection{Data Association}
\label{section_data_association}
Given the set $D^t$ of detections in key frame $t$, and the set $O^{t-1}$ of objects in the previous non-key frame $t-1$,
the data association process is divided into two steps.

Step $1$: assign the detections in $D^t$ to confirmed objects based on the IoU cost between bounding-boxes. 
Let $c_{i,j}^{iou}$ be the IoU cost between object $o_i^{t-1}$ and detection $d^t_j$
\begin{equation}
c_{i,j}^{iou} = 1 - {\rm IoU}(b_i^{t-1}, b_j^t),
\end{equation}
and they will not be associated with each other if $c_{i,j}^{iou}$ is greater than a threshold $\tau_{iou}$.

Step $2$: assign the unmatched detections to objects in tentative state as well as those unmatched objects in step 1 based on the appearance cost.
Let $c_{i,j}^{app}$ be the appearance cost between object $o_i^{t-1}$ and detection $d^t_j$
\begin{equation}
c_{i,j}^{app} = 1 - \max \limits_{\tau \in \{1, 2, ..., |F_i|\}} p_{i,j}^{\tau},
\end{equation}
where $p_{i,j}^{\tau}$ is the appearance affinity obtained by A-CNN.
And they will not be associated if $c_{i,j}^{app}$ is greater than a threshold $\tau_{app}$.

Suppose the detection $d^t_j$ is assigned to the object $o_i^{t-1}$,
the bounding-box of the $i$-th object in frame $t$ is succeeded from the bounding-box of $d_j^t$.
The detection's appearance feature $f_j^t$ is added to the feature set $F_i$.
The oldest feature will be abandoned if there are more than $l_{f}$ features in $F_i$.

\subsubsection{Object Management}
\label{section_object_management}
The objects are managed by transforming their states between the pre-defined three states $\{s_T, s_C, s_D\}$.
Particularly:
\begin{enumerate}
\item An unmatched detection is initialized as a tentative object, and it will be confirmed if its detection confidence is larger than a threshold $c_{s_T\rightarrow s_C}$.
\item A confirmed object is transformed to tentative state if it has not been associated with any detections for more than $l_{s_C \rightarrow s_T}$ consecutive key frames.
\item A tentative object will be confirmed if it has been associated with a detection for more than $l_{s_T \rightarrow s_C}$ consecutive key frames.
\item A tentative object will be deleted if it has not been associated with any detections for more than $l_{s_T \rightarrow s_D}$ consecutive key frames.
\item A deleted object remains at $s_D$ forever.
\end{enumerate}

\subsection{Tracking in Non-key Frames}
\label{subsection_tracking_cnn}
The tracking in non-key frames is much more straightforward.
The objects are directly propagated by T-CNN while maintaining their identities.
The appearance features and state of each object are not changed during propagation since the RGB images are not restored for detection and data association. 
Let $\hat{v}_i^t = (\hat{v}_{i,x}^t, \hat{v}_{i, y}^t, \hat{v}_{i,w}^t, \hat{v}_{i, h}^t)$ be the predicted velocity of the $i$-th object in frame $t$ based on the bounding-box in frame $t-1$:

\begin{equation}
\hat{v}_i^t = \mathcal{V}(b_i^{t-1}, \mathcal{N}_T(\mathfrak{f}^t)),
\label{equation_velocity}
\end{equation}
where $\mathcal{N}_T(\cdot)$ is the backbone CNN of T-CNN, 
and $\mathfrak{f}^t$ represents the input data for the non-key frame $t$.
$\mathcal{V}(\cdot, \cdot)$ is the velocity prediction function. 
As shown in Figure \ref{fig:tracking_cnn_overview},
$\mathcal{V}(\cdot, \cdot)$ is implemented by position-sensitive region-of-interest pooling layer (PSRoIPooling) proposed in R-FCN \cite{dai2016r}.
Then the bounding-boxes in frame $t$ can be predicted easily by $b_i^t = \mathcal{B}(\hat{v}_i^t, b_i^{t-1})$,
where $\mathcal{B}(\cdot, \cdot)$ is the bounding-box prediction function that defined as
\begin{equation}
\left\{
\begin{array}{lr}
    x^{t}_{i} = w^{t-1}_{i} \hat{v}^{t}_{i,x} + x^{t-1}_{i}, & \\ 
    y^{t}_{i} = h^{t-1}_{i} \hat{v}^{t}_{i,y} + y^{t-1}_{i},  & \\
    w^{t}_{i} = w^{t-1}_{i}  \exp(\hat{v}_{i,w}^{t-1}), & \\ 
    h^{t}_{i} = h^{t-1}_{i}  \exp(\hat{v}_{i,h}^{t-1}). &
\end{array}
\right.
\label{equation_bounding_box_prediction}
\end{equation}
Obviously, the identity is maintained for each object during the propagation process.

\noindent\textbf{Backbone CNN.}
The network used in T-CNN is much smaller than the network in detector, since the MVs and residuals only store the $changes$ between two frames.
Besides, a smaller network can reduce the computational cost.
The backbone CNN is modified from ResNet-18~\cite{he2016deep}.
Particularly, the last average pooling layer and fully connection layer are removed.
As a common practice \cite{dai2016r}, the effective stride of ResNet-18 is reduced from $32$ pixels to $16$ pixels, which increases the resolution of feature maps.
Then we can get three types of the modified ResNet-18 by changing the number of input channels in the first convolutional layer to 2, 3 and 5,
which are denoted as ResNet$_2$-18, ResNet$_3$-18 and ResNet$_5$-18, respectively.

\begin{figure}[tp]
\centering
    \includegraphics[width=1\linewidth]{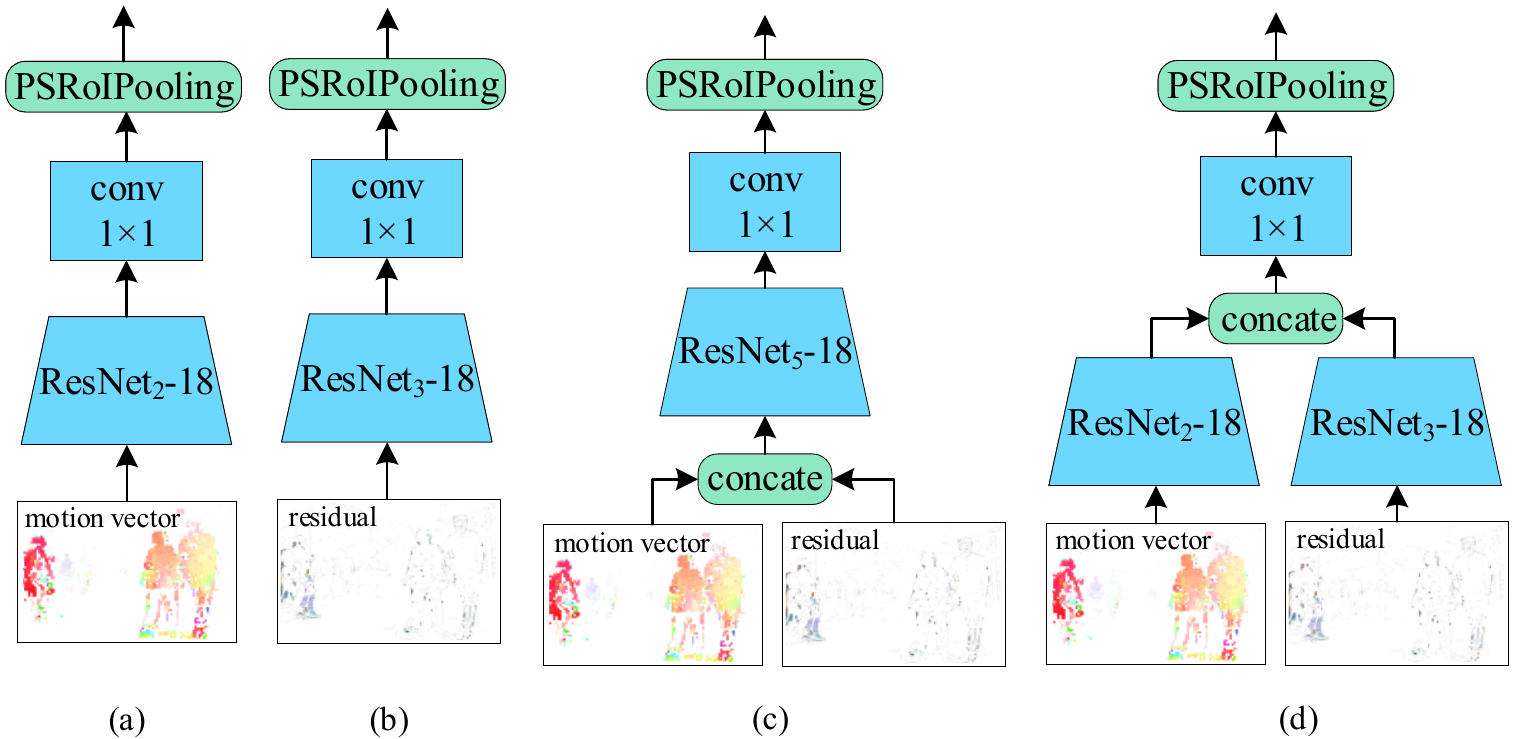}
    \caption{
    Four prototypes of T-CNN.
    (a): T-CNN$_{\rm mv}$, only MVs are fed into T-CNN.
    (b): T-CNN$_{\rm res}$, only residuals are fed into T-CNN.
    (C): T-CNN$_{\rm mv|res}$, MVs and residuals are concatenated together to be fed into T-CNN.
    (d): T-CNN$_{\rm mv||res}$, MVs and residuals are fed into ResNet$_2$-18 and resNet$_3$-18 respectively, then the outputs of these two CNNs are concatenated together.
    For all designs of T-CNN, the corresponding $1 \times 1$ convolutional layer (followed by a ReLU function) is used to produce a feature map with $4m^2$ channels.}
\label{fig:tracking_cnn_backbone}
\end{figure}

In order to explore the tracking ability of T-CNN, four T-CNNs are designed, as shown in Figure \ref{fig:tracking_cnn_backbone}:
\begin{itemize}
\item T-CNN$_{\rm mv}$: only MVs are used to predict the velocities. 
\item T-CNN$_{\rm res}$: only residuals are used to predicted the velocities. 
\item T-CNN$_{\rm mv|res}$: MVs and residuals are both used, but they are concatenated together firstly to be fed into T-CNN.
\item T-CNN$_{\rm mv||res}$: MVs and residuals are both used, and they are fed into their corresponding branches. Then the outputs of these two branches are concatenated together.
\end{itemize}
For all prototypes of T-CNN, the $1 \times 1$ convolutional layer (followed by a ReLU function) is used to produce a feature map with $4m^2$ channels.

\noindent\textbf{Training of T-CNN.}
The training of T-CNN is independent of the training of A-CNN and detector.
The reason is that detection and appearance feature extraction need to be performed on RGB images,
while T-CNN needs motion vectors and residuals to predict the velocities of objects.
Given the ground-truth bounding-boxes of the $i$-th object in frame $t-1$ and $t$,
the ground-truth velocity $v_i^t = (v_{i,x}^t, v_{i, y}^t, v_{i,w}^t, v_{i, h}^t)$ can be computed by
$ v_i^t = \mathcal{B}^{-1}(b_i^{t-1}, b_i^t)$,
where $\mathcal{B}^{-1}(\cdot,\cdot)$ is the inverse function of $\mathcal{B}(\cdot, \cdot)$.
The loss of T-CNN can be computed by
\begin{equation}
    L_T = \frac{1}{I} \sum^{I}_{i=1}\sum_{u \in \{x, y, w, h\}}{\rm smooth}_{L_1}(v_{i,u}^t - \hat{v}_{i,u}^{t}),
\label{equation_t_cnn_loss}
\end{equation}
in which
\begin{equation}
\rm{smooth_{L1}}(x)=\left\{
\begin{aligned}
0.5x^2 \quad  \quad if \ |x| < 1 \ \\
|x| - 0.5 \quad \  otherwise,
\end{aligned}
\right.
\label{equation_smooth_l1}
\end{equation}
is the function defined in \cite{girshick2015fast},
and $I$ is the number of objects that appear in frames $t$ and frame $t-1$.
The objects that only appear in one frame will not contribute to the training process.

\begin{algorithm}[tp]
\small
\caption{OTCD tracker}
\hspace*{0.02in} {\bf Input:}
a MPEG-4 video $v$ with $T$ frames, key frame scheduler $K$
\\ \hspace*{0.02in} {\bf Output:}
Trajectories of objects $B = \{\{b_i^t, id_i\}_{i=1}^{I^t}\}_{t=1}^{T}$
\begin{algorithmic}[1]
\State Initialization: $B \leftarrow \emptyset$, $O^0 \leftarrow \emptyset$, $t \leftarrow 1$
\While{$t \le T$}
    \State \scriptsize $/\star \texttt{track objects online} \star/$ \small
    \State $F^t \leftarrow \Call{Load}{v, t}$ \scriptsize $\qquad \qquad  \qquad  \qquad \qquad  \qquad//$ \texttt{load a frame} \small
    \If{$(t-1)$ mod $K = 0$ } \scriptsize $\qquad \qquad  \qquad \qquad //\  \texttt{a key frame}$ \small
        \State $F^t \leftarrow \Call{Restore}{F^t}$  \scriptsize $\qquad  \qquad  \quad \ \ //$ \texttt{restore RGB image} \small
        \State $D^t$ $\leftarrow \Call{R-fcn}{F^t}$ 
        \State $\Call{Association}{D^t, O^{t-1}}$ 
        \State $O^t \leftarrow \Call{Management}{D^t, O^{t-1}}$ 
    \Else \scriptsize $\qquad  \qquad  \qquad  \qquad  \qquad  \qquad  \qquad  \qquad  \qquad //\  \texttt{a non-key frame}$ \small
        \State $O^t \leftarrow \Call{Propagate}{O^{t-1}, F^t}$ 
    \EndIf
    \State \scriptsize $/\star \texttt{store the bounding-boxes and identities} \star/$ \small
    \State $B^t \leftarrow \emptyset$
    \For{each $o_i^t \in O^t$}
    \If{$s_i^t = s_C$}
        \State $B^t \leftarrow B^t \cup \{b_i^t, id_i\}$
    \EndIf
    \EndFor
    \State $B \leftarrow B \cup B^t$
\EndWhile
\State \Return $B$
\end{algorithmic}
\label{algorithm_overview}
\end{algorithm}

\subsection{Time Consumption Analysis}
Let $T_{det}$, $T_{ass}$, $T_{man}$ and $T_{pro}$ be the time consumption of detection, data association, object management and object propagation in each frame.
Compared to the per-frame approaches, the speedup factor $s$ of our tracker depends on the sparsity of key frames:
\begin{equation}
s = \frac{K(T_{det} + T_{ass} + T_{man})}{T_{det} + T_{ass} + T_{man} + (K-1)T_{pro}}.
\end{equation}
Generally, $T_{man} \ll T_{det}$ and $T_{man} \ll T_{ass}$.
In our implementation, $T_{pro} \approx \frac{T_{det} + T_{ass}}{10}$.
Then $s$ is about
\begin{equation}
s \approx \frac{10K}{K+9}.
\end{equation}
For example, our tracker is $2.5 \times$ faster approximately when $K = 3$.
The tracking method is shown in Algorithm \ref{algorithm_overview}.

\begin{table*}[tp]
\centering
\caption[width=1.\linewidth]{Tracking performance on validation set with different settings for data association.
Values in bold highlight the best results.}
\small
\setlength{\tabcolsep}{6pt}
\begin{tabular}{c|c|c|c|c|c}
\hline
\multicolumn{6}{c}{OTCD} \\ \hline
\diagbox{Metrics}{Settings}  &A-CNN$^-_{PS}$ &A-CNN &IoU &IoU+A-CNN &IoU $\rightarrow$ A-CNN \\ \hline
IDS$\downarrow$ &555 &348 &195 &167 &\textbf{141}\\ \hline
MOTA$\uparrow$ &32.0\% &33.2\% &35.0\% &35.4\% &\textbf{35.5\%}  \\ \hline
Hz$\uparrow$ &3.2 &2.3 &\textbf{16.5} &2.1 &15.7\\
\hline
\end{tabular}
\label{table:ablation_study_a-cnn}
\end{table*}

\section{EXPERIMENTS}
\label{section_experiments}
The proposed MOT tracker OTCD is implemented based on PyTorch library without optimization.
Evaluation is on a workstation with $2.6$ GHz CPU and Nvidia TITAN Xp GPU.

\subsection{Datasets}
\label{section_datasets}
We evaluate the proposed tracker on Citypersons~\cite{zhang2017citypersons}, 2DMOT2015~\cite{leal2015motchallenge}, MOT16~\cite{milan2016mot16} and MOT17~\cite{milan2016mot16}.
The sequences in MOT16 are the same with those in MOT17 but are provided with a less accurate ground-truth.
Each sequence in 2DMOT2015 and MOT17 is compressed into a MPEG-4 video, and all images in Citypersons are compressed into a MPEG-4 video.
All data used for training and testing is loaded from the compressed domain with the tool provided by \cite{wu2018compressed}.
The loaded MVs and residuals can be treated as special $images$ (2 and 3 channels respectively) which have the same resolution with the video.

Sequences in 2DMOT2015 and MOT17 are divided into three sets.
\emph{Testing set}: the sequences in MOT17 test split.
\emph{Validation set}: MOT17-09 and MOT17-10.
\emph{Training set}: the rest sequences in MOT17 as well as those sequences in 2DMOT2015 train split but not included in MOT17.

\subsection{Settings}
\label{section_settings}
The variable $m$ used in A-CNN and T-CNN is set to $7$.
The detections with confidence less than $0.95$ are abandoned.
The detection confidence threshold $c_{s_T \rightarrow s_C}$ is set to $0.99$.
The number of consecutive key frames $l_{s_T \rightarrow s_C}$,  $l_{s_C \rightarrow s_T}$, $l_{s_T \rightarrow s_D}$ are set to $3$, $2$, $10$ respectively.
And the number of historical appearance features $l_{f}$ is set to $24$.
The thresholds  $\tau_{iou}$ and $\tau_{app}$ are set to $0.3$ and $0.25$ respectively.

Citypersons and training set are used to train R-FCN and A-CNN.
The training samples for A-CNN are generated during training process.
Particularly, for each ground-truth box in one image, two positive and one negative samples are collected with $\ge 0.7$ and $\le 0.3$ IoU overlap ratios with this ground-truth box.
We also randomly choose a ground-truth box belonging to other objects as another negative sample.
The separate training of R-FCN and A-CNN are conducted with initial learning rate $10^{-3}$ and $10^{-4}$ respectively.
The joint training of R-FCN and A-CNN is divided into 3 phases.
Phase 1: train R-FCN  for $15$ epochs with initial learning rate $10^{-3}$. 
Phase 2: train A-CNN  for $5$ epochs with initial learning rate $10^{-4}$.
Phase 3: train A-CNN and R-FCN jointly with initial learning rate $10^{-4}$. 
Here, we set $\lambda$ to $1$.
T-CNN is trained on training set with initial learning rate $10^{-4}$. 
During all training process,
the batch size is set to $2$ and the learning rate decades every $8$ epoches with exponential decay rate $0.1$.
All models are optimized by stochastic gradient descent until they converge.

\begin{figure}[tp]
\centering
    \includegraphics[width=1.0\linewidth]{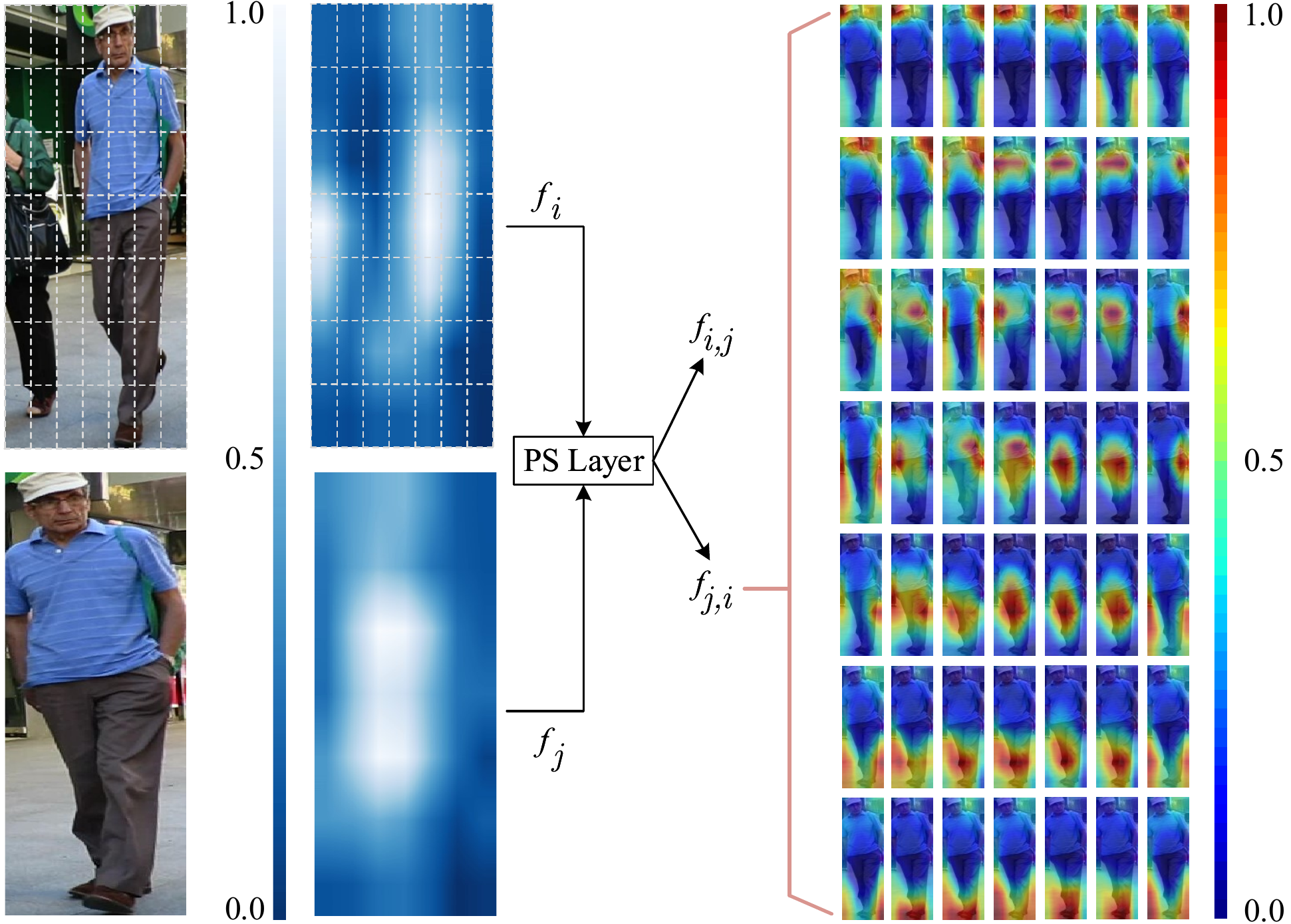}
    \caption{Qualitative results on PS layer.
    The feature maps are resized to the size of image patches.
    Left: two image patches (and their appearance features) need to be compared with.
    The pixel values plotted in the feature map are proportional to the $L_2$-norm of corresponding feature vectors along channel dimension.
    Right: two ($m^2$-channel) feature maps $f_{i,j}$ and $f_{j,i}$ that obtained by PS layer.
    Only $f_{j,i}$ is presented in detail.
    The top image patch is divided into $m \times m$ bins ($m = 7$ in our experiments),
    corresponding to the spatial size of the features that input to A-CNN.
    Each bin is compared with the bottom image patch, producing a single channel feature map in $f_{j,i}$.
    The similar two patterns are with a high similarity score.
    }
\label{fig:ps_layer_figure_demo}
\end{figure}

\subsection{Metrics}
\label{section_metrics}
\noindent\textbf{Trackers.}
We choose the following metrics to evaluate different trackers:
Multi-Object Tracking Accuracy (MOTA) \cite{bernardin2008evaluating},
Multi-Object Tracking Precision (MOTP) \cite{bernardin2008evaluating},
how often an object is identified by the same ID (IDF1) \cite{ristani2016performance},
Mostly Tracked objects (MT),
Mostly Lost objects (ML),
number of False Positives (FP),
number of False Negatives (FN),
number of Identity Switches (IDS) \cite{li2009learning},
number of Fragments (Frag),
and running speed (Hz).

\noindent\textbf{Detector.}
Except FP and FN, we choose additional metrics to evaluate different detectors:
Recall (Rcll),
Precision (Prcn),
Multiple Object Detection Accuracy (MODA) \cite{stiefelhagen2006clear}.

All metrics are evaluated by the toolkit provided by MOTChallenge benchmark \cite{leal2015motchallenge, milan2016mot16}.

\subsection{Components Analyses}
\label{section_ablation_study}
\subsubsection{Appearance CNN}
\label{section_ablation_study_a-cnn}
We first present the features produced by the PS layer in Figure \ref{fig:ps_layer_figure_demo}.
The PS layer actually divides one image patch into $m \times m$ bins,
and compares each bin with another image patch.
A high similarity score will be produced if the compared image patterns are similar.

In order to demonstrate the effectiveness of the two steps data association procedure introduced in section \ref{section_data_association} (denoted as IoU$\rightarrow$A-CNN),
an one step data association procedure that simultaneously take IoU cost and appearance cost into consideration are also conducted (denoted as IoU+A-CNN).
For the one step data association procedure, we use $c_{i,j}=\alpha c^{iou}_{i,j} + (1 - \alpha) c^{app}_{i,j}$ as the cost between object $o_i^{t-1}$ and detection $d^t_j$,
and they will not be associated with each other if $c_{i,j}$ is greater than the threshold $\tau = \alpha \tau^{iou} + (1 - \alpha) \tau^{app}$.
Several experiments are conducted by varying $\alpha$ from $0.1$ to $0.9$ with the interval $0.1$,
and the best MOTA is achieved on validation set when $\alpha = 0.5$,
which is our default setting for the one step data association procedure.
Note that $\alpha = 1$ and $\alpha = 0$ are two special cases of IoU+A-CNN which are denoted as IoU and A-CNN, respectively.
An A-CNN without PS layer (A-CNN$^-_{\rm PS}$) is also trained to further demonstrate the effectiveness of the PS layer in A-CNN.
The results are shown in Table \ref{table:ablation_study_a-cnn}.

When the appearance cost is only used, MOTA is improved by $1.2\%$ and IDS is greatly reduced by $37.3\%$ with the help of the PS layer at the cost of tracking speed dropped from 3.2 Hz to 2.3 Hz.
Favorable MOTA and IDS can be obtained when the IoU cost is only used.
This is due to the fact that the bounding boxes of one object in the adjacent frames may be much closer,
and the IoU cost is sufficient enough to associate them with each other.
The best tracking speed is also possessed by the case where the IoU cost is only used.
This is reasonable since appearance cost is no need to be computed, which is more computational expensive than IoU cost.
However, MOTA is still improved by $0.5\%$ and IDS is further reduced by $27.7\%$ when A-CNN is used in IoU$\rightarrow$A-CNN
with the little price of Hz dropped from 16.5 to 15.7.

Though MOTA is almost the same in IoU+A-CNN and IoU$\rightarrow$A-CNN, a better IDS and Hz are achieved by IoU$\rightarrow$A-CNN.
Here is the explanation: (1) Hz. Both IoU cost and appearance cost need to be computed for each object-detection pair in IoU+A-CNN.
While in IoU$\rightarrow$A-CNN, appearance cost (which is computational expensive) only needs to be computed for a small proportion of object-detection pairs.
(2) IDS and MOTA. IoU cost and appearance cost may not be always both reliable.
Simultaneously consideration of IoU cost and appearance cost will make them influence each other.

\begin{figure*}[tp]
\centering
    \includegraphics[width=0.99\linewidth]{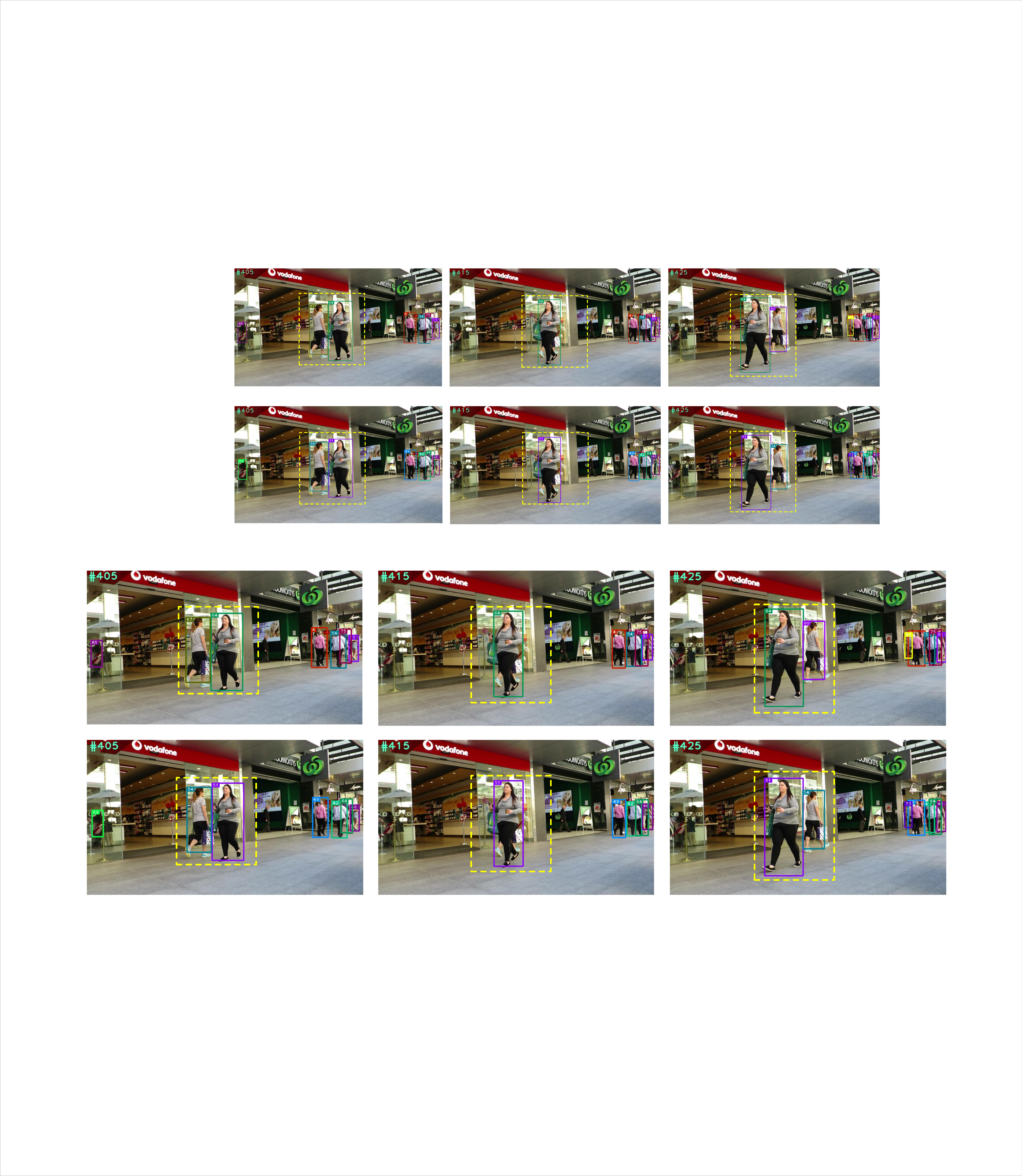}
    \caption{A case of occlusion in MOT17-09. Please pay attention to the two pedestrians that locates in yellow dashed boxes.
    The color of bounding-boxes and the numbers locate in top-left corner of the bounding-boxes denote the identities of corresponding pedestrian.
    From left to right: frame 405 (before occlusion occurs), frame 415 (when occlusion occurs), frame 425 (after occlusion occurs).
    Top row: data association is conducted based on the IoU cost only.
    The tracker fails to re-recognize the occluded pedestrian.
    Bottom row: data association is conducted based on the IoU cost and the appearance cost as described in section \ref{section_data_association}.
    The tracker re-recognizes the occluded pedestrian successfully.
    }
\label{fig:occlusion_happens}
\end{figure*}

A case of occlusion is also shown in Figure \ref{fig:occlusion_happens}.
When data association procedure is conducted based on the IoU cost only,
the tracker fails to re-recognize the occluded pedestrian.
However, when the data association procedure is conducted based on both IoU and appearance cost
as described in section \ref{section_data_association},
the tracker can re-recognize the occluded pedestrian successfully,
which demonstrates the effectiveness of the proposed A-CNN and the data association method.

\begin{table}[tp]
\centering
\caption[width=1.\linewidth]{The effect of joint training of A-CNN and detector on detection and tracking performance.
Tested on validation set.}
\small
\setlength{\tabcolsep}{3pt}
\begin{tabular}{c|c|c|c|c|c|c}

\hline
&joint training &Rcll$\uparrow$ &Prcn$\uparrow$ &FP$\downarrow$ &FN$\downarrow$ &MODA$\uparrow$ \\ \hline
\multirow{2}*{R-FCN}
& &55.1\% &85.4\% &1165 &5537 &45.6\% \\
&$\checkmark$ &\textbf{55.8\%} &\textbf{88.0\%} &\textbf{934} &\textbf{5448} &\textbf{48.2\%} \\ \hline

\multicolumn{7}{c}{} \\

\hline
 &joint training &MOTA$\uparrow$ &IDF1$\uparrow$ &FP$\downarrow$ &FN$\downarrow$  &Frag$\downarrow$ \\ \hline

\multirow{2}*{OTCD}
& &35.5 &22.9\% &1437 &\textbf{9759}  &362 \\
 &$\checkmark$ &\textbf{38.4\%} &\textbf{25.1\%} &\textbf{911} &9770  &\textbf{321} \\
\hline
\end{tabular}
\label{table:ablation_study_joint_training}
\end{table}

\subsubsection{Joint Training of Appearance CNN and Detector.}
\label{section_joint_training}
The effectiveness of joint training of A-CNN and detector is shown in Table \ref{table:ablation_study_joint_training}.
In terms of detection, all metrics are improved. Particularly, FP is greatly reduced about $19.8\%$, Prcn and MODA are both improved by more than $2\%$.
In terms of tracking, FP is greatly reduced about $36.6\%$, which leads to a higher overall tracking performance MOTA.
Furthermore, IDF1 is improved by $2.2\%$, which means the tracker can recognize an object with the same ID more often.

\begin{figure*}[tp]
\centering
\subfigure[]
{
    \begin{minipage}{0.47\linewidth}
    \centering
    \includegraphics[width=0.9\linewidth]{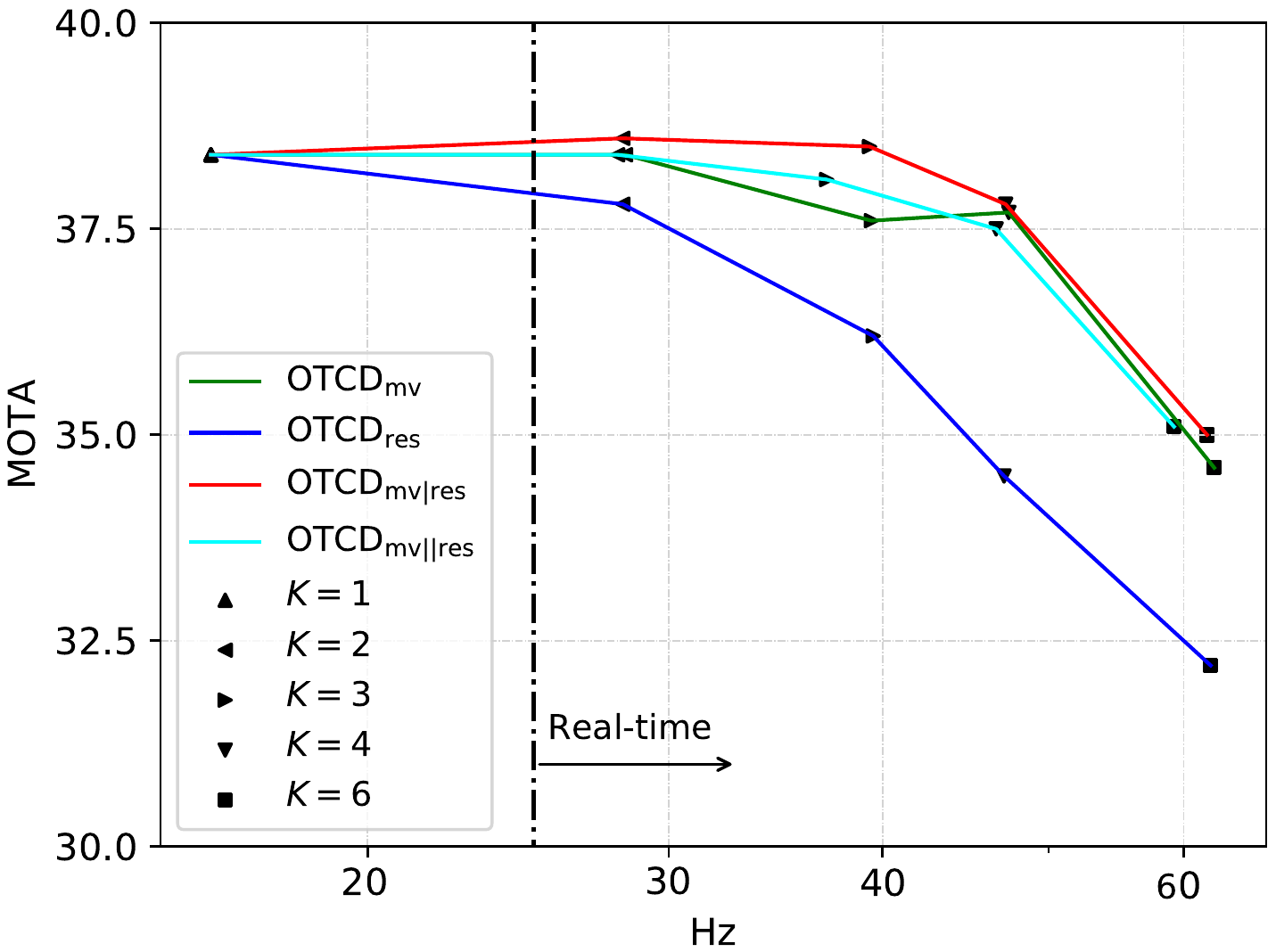}
    \end{minipage}
}
\hfill
\subfigure[]
{
    \begin{minipage}{0.47\linewidth}
    \centering
    \includegraphics[width=0.9\linewidth]{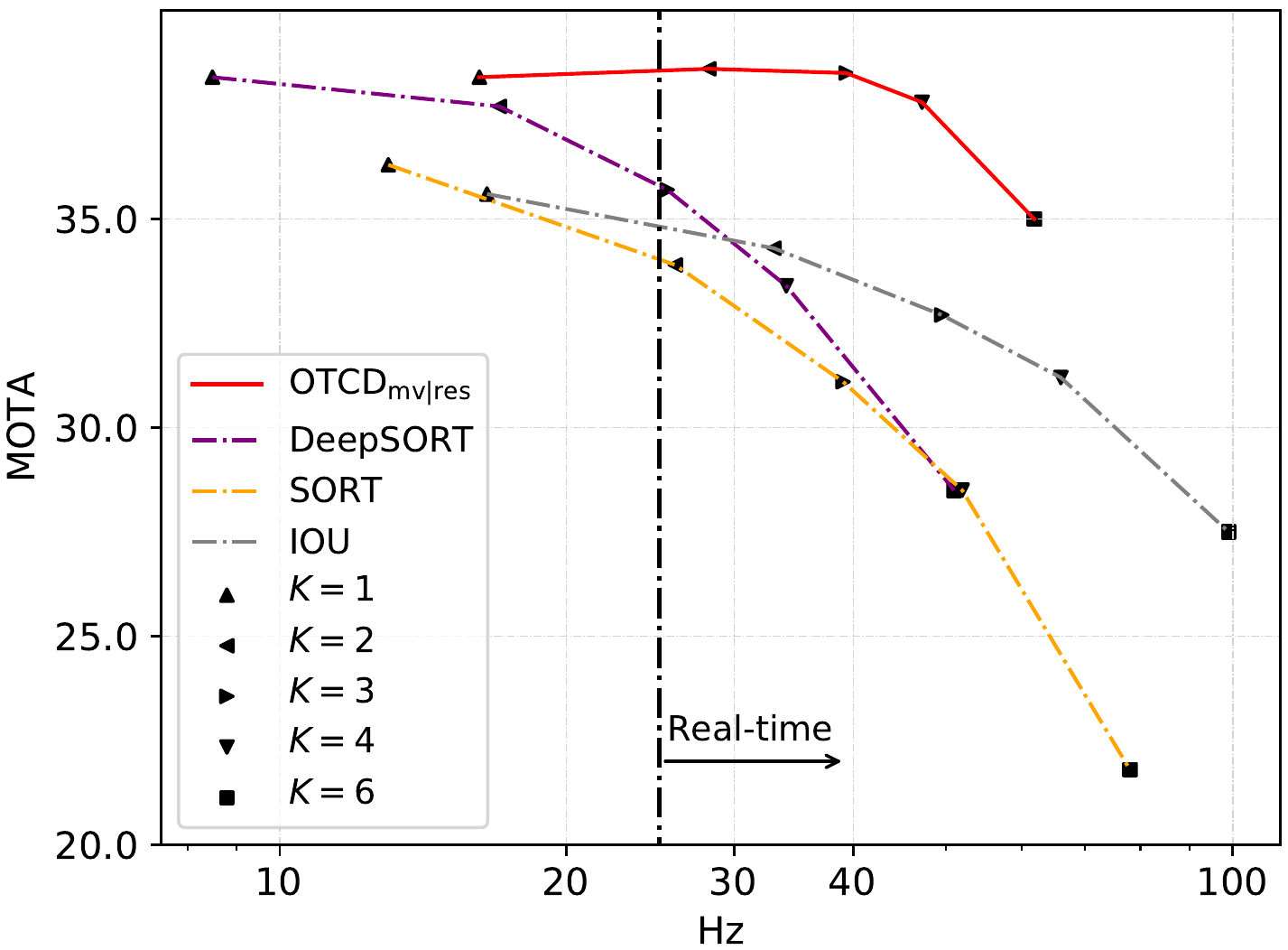}
    \end{minipage}
}
    \caption{
Performance-speed tradeoff with different $K$.
  The subscript of OTCD denotes the corresponding prototype of T-CNN.
  (a) Tracking results of OTCD with different prototypes of T-CNN.
  (b) Tracking results of our default tracker and other trackers.
  Tested on validation set.}
\label{fig_ablation_t_cnn}
\end{figure*}

\begin{table*}[tp]
\caption{Results of different trackers on MOTChallenge benchmark.
The subscript of OTCD is the value of $K$. Values in bold highlight the best results and Hz in {\color{blue}blue} highlight the real-time trackers.
Hz marked by $\S$ means detection time is considered.
Trackers marked by $\star$ and $\dag$ use the detections provided by \cite{yu2016poi} and ourselves, respectively.
While other trackers use the public detections.
}
\small
\setlength{\tabcolsep}{6pt}
\centering
\begin{tabular}{|c|c|c|c|c|c|c|c|c|c|c|c|}
\hline
benchmark &trackers  &MOTA$\uparrow$ &MOTP$\uparrow$ &IDF1$\uparrow$ &MT$\uparrow$ &ML$\downarrow$ &FP$\downarrow$ &FN$\downarrow$ &IDS$\downarrow$ &Frag$\downarrow$ &Hz$\uparrow$ \\
\hline
\hline
\multirow{15}*{MOT16}


&DeepSORT$^\star$\cite{wojke2017simple} &\textbf{61.4\%} &79.1\% &\textbf{62.2\%} &32.8\% &18.2\% &12852 &56668 &\textbf{781} &2008 &17.4 \\
&SORT$^\star$\cite{bewley2016simple} &59.8\% &\textbf{79.6\%} &53.8\% &25.4\% &22.7\% &\textbf{8698} &63245 &1423 &1835 &{\color{blue}\textbf{59.5}} \\
&MVint(LinearK)$^\star$ \cite{ujiie2018interpolation} &55.0\% &76.7\% &- &20.4\% &24.5\% &15766 &65297 & 1024 &\textbf{1594} &16.9  \\
&OTCD$_1^\star$ &60.2\% &79.1\% &55.1\% &\textbf{35.8\%} &\textbf{16.1\%} &14724 &55244 &2670 &2310 &12.6 \\

&OTCD$_3^\star$ &59.7\% &78.5\% &56.8\% &35.4\% &16.9\% &16882 &\textbf{55156} &1456 &1971 &{\color{blue}28.6} \\

&OTCD$_1^{\dag}$ &46.8\% &73.6\% &45.0\% &15.9\% &43.0\% &12914 &82777 &1236 &2666 &13.6$^\S$ \\

&OTCD$_3^{\dag}$ &46.6\% &74.2\% &46.4\% &15.5\% &45.2\% &12359 &84155 &783 &1984 &{\color{blue}30.3}$^\S$ \\

\cline{2-12}
&AMIR \cite{sadeghian2017tracking} &\textbf{47.2\%} &75.8\% &46.3\% &14.0\% &41.6\% &\textbf{2681} &92856 &774 &1675 &1.0 \\

&DMMOT \cite{zhu2018online} &46.1\% &73.8\% &\textbf{54.8\%} &\textbf{17.4\%} &42.7\% &7909 &89874 &532 &1616 &0.3 \\

&STAM16 \cite{chu2017online} &46.0\% &74.9\% &50.0\% &14.6\% &43.6\% &6895 &91117 &\textbf{473} &1422 &0.2 \\

&MTDF \cite{fu2019multi} &45.7\% &72.6\% &40.1\% &14.1\% &\textbf{36.4\%} &12018 &\textbf{84970} &1987 &3377 &1.5 \\


&PHD\_GSDL16 \cite{fu2018particle} &41.0\% &\textbf{75.9\%} &43.1\% &11.3\% &41.5\% &6498 &99257 &1810 &3650 &8.3 \\

&AM\_ADM \cite{lee2018learning} &40.1\% &75.4\% &43.8\% &7.1\% &46.2\% &8503 &99891 &789 &1736 &5.8 \\

&OTCD$_1$ &44.4\% &75.4\% &45.6\% &11.6\% &47.6\% &5759 &97927 &759 &1787 &17.6 \\

&OTCD$_3$ &42.4\% &75.7\% &46.4\% &11.1\% &50.1\% &8318 &96177 &570 &\textbf{1299} &{\color{blue}\textbf{31.7}} \\

\hline
\hline

\multirow{11}*{MOT17}
&MTDF17 \cite{fu2019multi} &\textbf{49.6\%} &75.5\% &45.2\% &18.9\% &\textbf{33.1\%} &37124 &\textbf{241768} &5567 &9260 &1.2 \\
&HAM\_SADF17 \cite{yoon2018online} &48.3\% &\textbf{77.2\%} &51.1\% &17.1\% &41.7\% &20967 &269038 &\textbf{1871} &\textbf{3020} &5.0 \\

&DMAN \cite{zhu2018online} &48.2\% &75.7\% &\textbf{55.7\%} &\textbf{19.3\%} &38.3\% &26218 &263608 &2194 &5378 &0.3\\

&AM\_ADM17 \cite{lee2018learning} &48.1\% &76.7\% &52.1\% &13.4\% &39.7\% &25061 &265495 &2214 &5027 &5.7 \\
&PHD\_GSDL17 \cite{fu2018particle} &48.0\% &\textbf{77.2\%} &49.6\% &17.1\% &35.6\% &23199 &265954 &3998 &8886 &6.7 \\

&FPSN \cite{lee2019multiple} &44.9\% &76.6\% &48.4\% &16.5\% &35.8\% &33757 &269952 &7136 &14491 &10.1 \\
&EAMTT \cite{sanchez2016online} &42.6\% &76.0\% &41.8\% &12.7\% &42.7\% &30711 &288474 &4488 &5720 &1.4 \\

&GM\_KCF \cite{kutschbach2017sequential} &39.6\% &74.5\% &36.6\% &8.8\% &43.3\% &50903 &284228 &5811 &7414 &3.3 \\
&GM\_PHD \cite{eiselein2012real} &36.4\% &76.2\% &33.9\% &4.1\% &57.3\% &23723 &330767 &4607 &11317 &{\color{blue}\textbf{38.4}}\\
&OTCD$_1$  &48.6\% &76.9\% &47.9\% &16.2\% &41.2\% &\textbf{18499} &268204 &3502 &5588 &15.5\\

&OTCD$_3$  &46.9\% &76.8\% &49.0\% &14.9\% &44.0\% &25204 &272153 &2154 &4072 &{\color{blue}33.4}   \\

\hline
\end{tabular}
\label{table:results on benchmark}
\end{table*}

\subsubsection{Tracking CNN}
\label{section_ablation_study_t-cnn}
We use T-CNN to speedup OTCD by varying $K \in \{1, 2, 3, 4, 6\}$.
The four prototypes of T-CNN are firstly evaluated,
as shown in Figure \ref{fig_ablation_t_cnn} (a).
The subscript of OTCD denotes the corresponding prototype of T-CNN.
Among our four trackers, OTCD$_{\rm res}$ performs the worst since no motion information is provided.
Compared with OTCD$_{\rm mv|res}$ and OTCD$_{\rm mv||res}$,  OTCD$_{\rm mv}$ is less stable,
we argue this to the absence of residuals.
Based on the above analysis, we can find that MVs and residuals complement each other.
Compared with OTCD$_{\rm mv||res}$,
OTCD$_{\rm mv|res}$ has a superior tracking capability.
What's more, the amount of parameters in T-CNN$_{\rm mv|res}$ is much smaller than that in T-CNN$_{\rm mv||res}$.
We choose OTCD$_{\rm mv|res}$ as our default tracker.

To demonstrate the effectiveness of the proposed T-CNN, we compare our method with some other trackers,
including DeepSORT \cite{wojke2017simple}, SORT \cite{bewley2016simple} and IOU \cite{bochinski2017high}.
The results are summarized in Figure \ref{fig_ablation_t_cnn} (b).
The bounding-boxes of objects in the non-key frames are predicted by Kalman filter in DeepSORT and SORT, while they are copied from previous frame in IOU.
For all trackers, the detection time (about $60$ ms) is considered and only detections in the key frames are provided for a fair comparison.
However, the time consumption in the non-key frames are not considered for DeepSORT, SORT and IOU.
As we can see, the tracking speed is limited due to the detection time consumption when $K=1$.
But all trackers achieve significant speedup with the descent tracking accuracy drop when $K > 1$.
Thanks to the powerful tracking capability of T-CNN, 
our tracker possesses the slowest performance decline when compared with other trackers.
For example, OTCD$_{\rm mv|res}$ is accelerated from $15.8$ Hz to $46.0$ Hz at the cost of accuracy drop from $38.4\%$ to $37.8\%$,
while DeepSORT is accelerated from $8.5$ Hz to $34.0$ Hz at the cost of accuracy drop from $38.4\%$ to $33.4\%$.

The effective performance gains brought by T-CNN does not hold when $K$ increases.
Reasons are as follows:
(1) Tracking accuracy. The birth and death of objects are not handled in the non-key frames, but objects may disappear or reappear during these frames,
which leads to a poorer tracking performance when $K$ increases.
(2) Tracking speed. Since we track objects without detection or data association on the non-key frames, the bounding-boxes of tracked objects may be imperfect,
which results in more imperfect bounding-boxes in the following non-key frames.
So more confirmed objects will not be associated with detections based on the IoU cost when the next one key frame arrives.
Hence, more detections need to be assigned to objects based on the appearance cost, which is more time-consuming than the IoU cost.
We choose $K=3$ for the balance of performance and speed.

\subsection{MOTChallenge Benchmark}
\label{subsection_motchallenge_benchmark}
We compare the proposed tracker OTCD with other online state-of-the-art trackers in Table \ref{table:results on benchmark} on MOT16 and MOT17 test splits.
During this evaluation, the detector is used for feature extraction, which means except for the backbone network of the detector, other parts of the detector are not used.
The metrics of MVint(LinearK) are accessed from \cite{ujiie2018interpolation}, while others are accessed from MOTChallenge leaderboards.

\noindent\textbf{MOT16.}
We first compare OTCD with some trackers that track objects based on the detections provided by \cite{yu2016poi}.
Note that the tracking accuracy MOTA and speed Hz of DeepSORT is superior to those of OTCD$^\star_1$,
which is not the result reflected in Figure \ref{fig_ablation_t_cnn} (b),
the reasons are as follows:
(1) MOTA. The detections used in Figure \ref{fig_ablation_t_cnn} (b) and Table \ref{table:results on benchmark} are different.
The quality of our own detections is inferior to the detections provided by \cite{yu2016poi}.
And a poor quality of detections has a negative impact on trackers. 
(2) Hz. Detection time consumption is considered for DeepSORT, SORT and OTCD in Figure \ref{fig_ablation_t_cnn} (b),
while it is not considered in Table \ref{table:results on benchmark}.

Both MVint(LinearK) and OTCD$^\star_3$ are in the compressed domain, but
OTCD$^\star_3$ achieves a better performance than MVint(LinearK) in all metrics except FP, IDS and Frag.
Particularly, the tracking speed of OTCD$^\star_3$ is about $1.7\times$ faster than MVint(LinearK), while possessing a $4.7\%$ higher MOTA.
The tracking results of OTCD based on our own detections are also provided.
Note that there is a big gap of MOTA between OTCD$^\star_1$ and OTCD$^\dag_1$, which means the quality of detections has a significant impact on the tracking performance.

As for the public detections, OTCD achieves the fastest tracking speed among all trackers while maintaining a satisfying tracking accuracy. 
For example, the tracking speed of OTCD$^\star_1$ is $17\times$ faster than
AMIR, which is the state-of-the-art method.
Furthermore, OTCD$^\star_3$ achieves the best performance in Frag.

Interestingly, when $K$ is increased from $1$ to $3$, both IDS and Frag are greatly reduced. 
This is reasonable since objects need to be recognized with a lower frequency when data association is only performed in sparse key frames.

\noindent\textbf{MOT17.} Overall, the tracking performance of OTCD is comparable with other trackers.
Particularly, OTCD$_1$ performs the best in FP, and OTCD$_3$ takes the second place in Hz.
Compared with MTDF17, which achieves the best in MOTA, our trackers run more than $10\times$ (OTCD$_1$) and $25\times$ (OTCD$_3$) faster
but only with $1.0\%$ (OTCD$_1$) and $2.7\%$ (OTCD$_3$) performance degradation in MOTA.
Compared with GM\_PHD, which possesses the best tracking speed (38.4 Hz), OTCD$_3$ tracks object at a lower speed (33.4 Hz), but with a 10.5\% higher MOTA.

\section{Conclusion}
\label{section_conclusion}
In this paper, we propose an online MOT tracker OTCD in compressed domain.
The RGB images are restored in the key frames for detection and data association,
while the MVs and residuals are directly fed into a tracking CNN to propagate objects through non-key frames,
which can accelerate our tracker significantly. 
Furthermore, an appearance CNN which shares features with detector is introduced to assist data association, and it can be trained  with detector jointly.
Experimental results on MOTChallenge benchmark demonstrate the effectiveness of appearance CNN and tracking CNN, as well as the joint training of appearance CNN and detector.

\ifCLASSOPTIONcompsoc
  \section*{Acknowledgments}
\else
  \section*{Acknowledgment}
\fi
This work is supported by the National Natural Science Foundation of China (Grant No. 61371192),
the Key Laboratory Foundation of the Chinese Academy of Sciences (CXJJ-17S044) and the Fundamental Research
Funds for the Central Universities (WK2100330002).

\ifCLASSOPTIONcaptionsoff
  \newpage
\fi



%

\bibliographystyle{IEEEtran}
\bibliography{cite}

\begin{thebibliography}{10}
\providecommand{\url}[1]{#1}
\csname url@samestyle\endcsname
\providecommand{\newblock}{\relax}
\providecommand{\bibinfo}[2]{#2}
\providecommand{\BIBentrySTDinterwordspacing}{\spaceskip=0pt\relax}
\providecommand{\BIBentryALTinterwordstretchfactor}{4}
\providecommand{\BIBentryALTinterwordspacing}{\spaceskip=\fontdimen2\font plus
\BIBentryALTinterwordstretchfactor\fontdimen3\font minus
  \fontdimen4\font\relax}
\providecommand{\BIBforeignlanguage}[2]{{%
\expandafter\ifx\csname l@#1\endcsname\relax
\typeout{** WARNING: IEEEtran.bst: No hyphenation pattern has been}%
\typeout{** loaded for the language `#1'. Using the pattern for}%
\typeout{** the default language instead.}%
\else
\language=\csname l@#1\endcsname
\fi
#2}}
\providecommand{\BIBdecl}{\relax}
\BIBdecl

\bibitem{zhang2008global}
L.~Zhang, Y.~Li, and R.~Nevatia, ``Global data association for multi-object
  tracking using network flows,'' in \emph{Computer Vision and Pattern
  Recognition, 2008. CVPR 2008. IEEE Conference on}.\hskip 1em plus 0.5em minus
  0.4em\relax IEEE, 2008, pp. 1--8.

\bibitem{tang2017multiple}
S.~Tang, M.~Andriluka, B.~Andres, and B.~Schiele, ``Multiple people tracking by
  lifted multicut and person reidentification,'' in \emph{Proceedings of the
  IEEE Conference on Computer Vision and Pattern Recognition}, 2017, pp.
  3539--3548.

\bibitem{yang2012online}
B.~Yang and R.~Nevatia, ``An online learned crf model for multi-target
  tracking,'' in \emph{Computer Vision and Pattern Recognition (CVPR), 2012
  IEEE Conference on}.\hskip 1em plus 0.5em minus 0.4em\relax IEEE, 2012, pp.
  2034--2041.

\bibitem{bae2018confidence}
S.-H. Bae and K.-J. Yoon, ``Confidence-based data association and
  discriminative deep appearance learning for robust online multi-object
  tracking,'' \emph{IEEE transactions on pattern analysis and machine
  intelligence}, vol.~40, no.~3, pp. 595--610, 2018.

\bibitem{zhu2018online}
J.~Zhu, H.~Yang, N.~Liu, M.~Kim, W.~Zhang, and M.-H. Yang, ``Online
  multi-object tracking with dual matching attention networks,'' in
  \emph{Proceedings of the European Conference on Computer Vision (ECCV)},
  2018, pp. 366--382.

\bibitem{wojke2017simple}
N.~Wojke, A.~Bewley, and D.~Paulus, ``Simple online and realtime tracking with
  a deep association metric,'' in \emph{Image Processing (ICIP), 2017 IEEE
  International Conference on}.\hskip 1em plus 0.5em minus 0.4em\relax IEEE,
  2017, pp. 3645--3649.

\bibitem{chu2017online}
Q.~Chu, W.~Ouyang, H.~Li, X.~Wang, B.~Liu, and N.~Yu, ``Online multi-object
  tracking using cnn-based single object tracker with spatial-temporal
  attention mechanism,'' in \emph{2017 IEEE International Conference on
  Computer Vision (ICCV).(Oct 2017)}, 2017, pp. 4846--4855.

\bibitem{ujiie2018interpolation}
T.~Ujiie, M.~Hiromoto, and T.~Sato, ``Interpolation-based object detection
  using motion vectors for embedded real-time tracking systems,'' in
  \emph{Proceedings of the IEEE Conference on Computer Vision and Pattern
  Recognition Workshops}, 2018, pp. 616--624.

\bibitem{yoon2018online}
Y.-c. Yoon, A.~Boragule, Y.-m. Song, K.~Yoon, and M.~Jeon, ``Online
  multi-object tracking with historical appearance matching and scene adaptive
  detection filtering,'' in \emph{2018 15th IEEE International Conference on
  Advanced Video and Signal Based Surveillance (AVSS)}.\hskip 1em plus 0.5em
  minus 0.4em\relax IEEE, 2018, pp. 1--6.

\bibitem{fu2018particle}
Z.~Fu, P.~Feng, F.~Angelini, J.~Chambers, and S.~M. Naqvi, ``Particle phd
  filter based multiple human tracking using online group-structured dictionary
  learning,'' \emph{IEEE Access}, vol.~6, pp. 14\,764--14\,778, 2018.

\bibitem{sanchez2016online}
R.~Sanchez-Matilla, F.~Poiesi, and A.~Cavallaro, ``Online multi-target tracking
  with strong and weak detections,'' in \emph{European Conference on Computer
  Vision}.\hskip 1em plus 0.5em minus 0.4em\relax Springer, 2016, pp. 84--99.

\bibitem{kutschbach2017sequential}
T.~Kutschbach, E.~Bochinski, V.~Eiselein, and T.~Sikora, ``Sequential sensor
  fusion combining probability hypothesis density and kernelized correlation
  filters for multi-object tracking in video data,'' in \emph{2017 14th IEEE
  International Conference on Advanced Video and Signal Based Surveillance
  (AVSS)}.\hskip 1em plus 0.5em minus 0.4em\relax IEEE, 2017, pp. 1--5.

\bibitem{yu2016poi}
F.~Yu, W.~Li, Q.~Li, Y.~Liu, X.~Shi, and J.~Yan, ``Poi: Multiple object
  tracking with high performance detection and appearance feature,'' in
  \emph{European Conference on Computer Vision}.\hskip 1em plus 0.5em minus
  0.4em\relax Springer, 2016, pp. 36--42.

\bibitem{eiselein2012real}
V.~Eiselein, D.~Arp, M.~P{\"a}tzold, and T.~Sikora, ``Real-time multi-human
  tracking using a probability hypothesis density filter and multiple
  detectors,'' in \emph{Advanced Video and Signal-Based Surveillance (AVSS),
  2012 IEEE Ninth International Conference on}.\hskip 1em plus 0.5em minus
  0.4em\relax IEEE, 2012, pp. 325--330.

\bibitem{kieritz2016online}
H.~Kieritz, S.~Becker, W.~H{\"u}bner, and M.~Arens, ``Online multi-person
  tracking using integral channel features,'' in \emph{Advanced Video and
  Signal Based Surveillance (AVSS), 2016 13th IEEE International Conference
  on}.\hskip 1em plus 0.5em minus 0.4em\relax IEEE, 2016, pp. 122--130.

\bibitem{bernardin2008evaluating}
K.~Bernardin and R.~Stiefelhagen, ``Evaluating multiple object tracking
  performance: the clear mot metrics,'' \emph{Journal on Image and Video
  Processing}, vol. 2008, p.~1, 2008.

\bibitem{alvar2018mv}
S.~R. Alvar and I.~V. Baji{\'c}, ``Mv-yolo: motion vector-aided tracking by
  semantic object detection,'' in \emph{2018 IEEE 20th International Workshop
  on Multimedia Signal Processing (MMSP)}.\hskip 1em plus 0.5em minus
  0.4em\relax IEEE, 2018, pp. 1--5.

\bibitem{wang2019detection}
H.~L. Shiyao~Wang, ``Fast object detection in compressed video,'' \emph{arXiv
  preprint arXiv:1811.11057}, 2019.

\bibitem{wu2018compressed}
C.-Y. Wu, M.~Zaheer, H.~Hu, R.~Manmatha, A.~J. Smola, and
  P.~Kr{\"a}henb{\"u}hl, ``Compressed video action recognition,'' in
  \emph{Proceedings of the IEEE Conference on Computer Vision and Pattern
  Recognition}, 2018, pp. 6026--6035.

\bibitem{kuo2010multi}
C.-H. Kuo, C.~Huang, and R.~Nevatia, ``Multi-target tracking by on-line learned
  discriminative appearance models,'' in \emph{Computer Vision and Pattern
  Recognition (CVPR), 2010 IEEE Conference on}.\hskip 1em plus 0.5em minus
  0.4em\relax IEEE, 2010, pp. 685--692.

\bibitem{wang2014tracklet}
B.~Wang, G.~Wang, K.~Luk~Chan, and L.~Wang, ``Tracklet association with online
  target-specific metric learning,'' in \emph{Proceedings of the IEEE
  Conference on Computer Vision and Pattern Recognition}, 2014, pp. 1234--1241.

\bibitem{union1994generic}
I.~T. Union-Telecommun, ``Generic coding of moving pictures and associated
  audio information-part 2: Video,'' \emph{Int. Standards Org./Int.
  Electrotech. Comm.(ISO/IEC) JTC 1, Rec. H. 262 and ISO/IEC 13 818-2 (MPEG-2
  Video)}, 1994.

\bibitem{wiegand2003draft}
T.~Wiegand, ``Draft itu-t recommendation and final draft international standard
  of joint video specification (itu-t rec. h. 264| iso/iec 14496-10 avc),''
  \emph{JVT-G050}, 2003.

\bibitem{he2017mask}
K.~He, G.~Gkioxari, P.~Doll{\'a}r, and R.~Girshick, ``Mask r-cnn,'' in
  \emph{Proceedings of the IEEE international conference on computer vision},
  2017, pp. 2961--2969.

\bibitem{dai2016r}
J.~Dai, Y.~Li, K.~He, and J.~Sun, ``R-fcn: Object detection via region-based
  fully convolutional networks,'' in \emph{Advances in neural information
  processing systems}, 2016, pp. 379--387.

\bibitem{he2016deep}
K.~He, X.~Zhang, S.~Ren, and J.~Sun, ``Deep residual learning for image
  recognition,'' in \emph{Proceedings of the IEEE conference on computer vision
  and pattern recognition}, 2016, pp. 770--778.

\bibitem{girshick2015fast}
R.~Girshick, ``Fast r-cnn,'' in \emph{Proceedings of the IEEE international
  conference on computer vision}, 2015, pp. 1440--1448.

\bibitem{zhang2017citypersons}
S.~Zhang, R.~Benenson, and B.~Schiele, ``Citypersons: A diverse dataset for
  pedestrian detection,'' in \emph{The IEEE Conference on Computer Vision and
  Pattern Recognition (CVPR)}, vol.~1, no.~2, 2017, p.~3.

\bibitem{leal2015motchallenge}
L.~Leal-Taix{\'e}, A.~Milan, I.~Reid, S.~Roth, and K.~Schindler, ``Motchallenge
  2015: Towards a benchmark for multi-target tracking,'' \emph{arXiv preprint
  arXiv:1504.01942}, 2015.

\bibitem{milan2016mot16}
A.~Milan, L.~Leal-Taix{\'e}, I.~Reid, S.~Roth, and K.~Schindler, ``Mot16: A
  benchmark for multi-object tracking,'' \emph{arXiv preprint
  arXiv:1603.00831}, 2016.

\bibitem{ristani2016performance}
E.~Ristani, F.~Solera, R.~Zou, R.~Cucchiara, and C.~Tomasi, ``Performance
  measures and a data set for multi-target, multi-camera tracking,'' in
  \emph{European Conference on Computer Vision}.\hskip 1em plus 0.5em minus
  0.4em\relax Springer, 2016, pp. 17--35.

\bibitem{li2009learning}
Y.~Li, C.~Huang, and R.~Nevatia, ``Learning to associate: Hybridboosted
  multi-target tracker for crowded scene,'' in \emph{2009 IEEE Conference on
  Computer Vision and Pattern Recognition}.\hskip 1em plus 0.5em minus
  0.4em\relax IEEE, 2009, pp. 2953--2960.

\bibitem{stiefelhagen2006clear}
R.~Stiefelhagen, K.~Bernardin, R.~Bowers, J.~Garofolo, D.~Mostefa, and
  P.~Soundararajan, ``The clear 2006 evaluation,'' in \emph{International
  evaluation workshop on classification of events, activities and
  relationships}.\hskip 1em plus 0.5em minus 0.4em\relax Springer, 2006, pp.
  1--44.

\bibitem{bewley2016simple}
A.~Bewley, Z.~Ge, L.~Ott, F.~Ramos, and B.~Upcroft, ``Simple online and
  realtime tracking,'' in \emph{Image Processing (ICIP), 2016 IEEE
  International Conference on}.\hskip 1em plus 0.5em minus 0.4em\relax IEEE,
  2016, pp. 3464--3468.

\bibitem{sadeghian2017tracking}
A.~Sadeghian, A.~Alahi, and S.~Savarese, ``Tracking the untrackable: Learning
  to track multiple cues with long-term dependencies,'' \emph{arXiv preprint
  arXiv:1701.01909}, vol.~4, no.~5, p.~6, 2017.

\bibitem{fu2019multi}
Z.~Fu, F.~Angelini, J.~Chambers, and S.~M. Naqvi, ``Multi-level cooperative
  fusion of gm-phd filters for online multiple human tracking,'' \emph{IEEE
  Transactions on Multimedia}, 2019.

\bibitem{lee2018learning}
S.-H. Lee, M.-Y. Kim, and S.-H. Bae, ``Learning discriminative appearance
  models for online multi-object tracking with appearance discriminability
  measures,'' \emph{IEEE Access}, vol.~6, pp. 67\,316--67\,328, 2018.

\bibitem{lee2019multiple}
S.~Lee and E.~Kim, ``Multiple object tracking via feature pyramid siamese
  networks,'' \emph{IEEE Access}, vol.~7, pp. 8181--8194, 2019.

\bibitem{bochinski2017high}
E.~Bochinski, V.~Eiselein, and T.~Sikora, ``High-speed tracking-by-detection
  without using image information,'' in \emph{Advanced Video and Signal Based
  Surveillance (AVSS), 2017 14th IEEE International Conference on}.\hskip 1em
  plus 0.5em minus 0.4em\relax IEEE, 2017, pp. 1--6.

\end{thebibliography}

\end{document}